%% file: main.tex
\documentclass[sigconf]{acmart}

\AtBeginDocument{%
  \providecommand\BibTeX{{%
    \normalfont B\kern-0.5em{\scshape i\kern-0.25em b}\kern-0.8em\TeX}}}

\setcopyright{rightsretained}

\makeatletter
\renewcommand\@formatdoi[1]{\ignorespaces}
\renewcommand\@acmBooktitle[1]{KDD Workshop on Machine Learning in Finance (KDD MLF '20)}
\makeatother

\acmConference[KDD MLF '20]{KDD '20: ACM Workshop Machine Learning in Finance}{August 23--27, 2020}{San Diego, CA}
\acmISBN{}

\usepackage{times}
\usepackage{soul}
\usepackage{url}
\usepackage{amsmath}
\usepackage{amsthm}
\usepackage{booktabs}
\usepackage{algorithm}
\usepackage{algorithmic}
\usepackage{microtype}
\usepackage[utf8]{inputenc}
\newtheorem{definition}{Definition}

\title{Predicting Account Receivables with Machine Learning}

\author[Ana P. Appel]{Ana Paula Appel}
\email{apappel@br.ibm.com}
\authornotemark[1]
\affiliation{%
  \institution{IBM Research}
  \city{São Paulo}
  \state{Brazil}
}
\author[Gabriel L. Malfatti]{Gabriel Louzada Malfatti}
\email{gabriel.malfatti@ibm.com}
\authornotemark[1]
\affiliation{%
  \institution{IBM Research}
  \city{São Paulo}
  \state{Brazil}
}
\author[Renato Luiz de F. Cunha]{Renato Luiz de Freitas Cunha}
\email{renatoc@br.ibm.com}
\authornotemark[1]
\affiliation{%
  \institution{IBM Research}
  \city{São Paulo}
  \state{Brazil}
}
\author{Bruno Lima}
\email{blima@br.ibm.com}
\authornotemark[1]
\affiliation{%
  \institution{IBM Research}
  \city{São Paulo}
  \state{Brazil}
}
\author{Rogerio de Paula}
\email{ropaula@br.ibm.com}
\authornotemark[1]
\affiliation{%
  \institution{IBM Research}
  \city{São Paulo}
  \state{Brazil}
}

\makeatletter
\g@addto@macro\@floatboxreset\centering
\makeatother

\begin{document}

\begin{abstract}

Being able to predict when invoices will be paid is valuable in multiple industries and supports decision-making processes in most financial workflows.
However, due to the complexity of data related to invoices and the fact that the decision-making process is not registered in the accounts receivable system, performing this prediction becomes a challenge.
In this paper, we present a prototype able to support collectors in predicting the payment of invoices. This prototype is part of a solution developed in partnership with a multinational bank and
it has reached up to 81\% of prediction accuracy, which improved the prioritization of customers and supported the daily work of collectors. Our simulations show that adoption of our model to prioritize the work o collectors saves up to $\approx$1.75 million dollars per month.
The methodology and results presented in this paper will allow researchers and practitioners in dealing with the problem of invoice payment prediction, providing insights and examples of how to tackle issues present in \emph{real data}.
\end{abstract}

%


\maketitle

\input{introduction}

\section{Problem Definition} \label{sec:problem}

In AR collection, the ability of monitoring and collecting payments enables the prediction of payment behavior. Firms often use various types of metrics to measure the performance of the collection process. One example is the average number of days overdue. Particularly in our case, the client is interested mainly in knowing the probability that an invoice will be paid late or on time to then be able to better prioritize the collection.  

The problem of predicting an invoice payment is a typical classification problem using supervised learning~\cite{James:2014} where, given the original client dataset, we need to extract invoices' features to be able to characterize each invoice with respect to labeled classes, building then a machine learning model to perform classification of new invoices.

In a more formal way, we can define our problem as follows:

\begin{definition}
Let $\mathcal{M} = {I,Y}$ be a set of pairs of invoices and their respective classes. Element $M_m$ is represented by the pair $\langle I_m, Y_m \rangle$, with $I_m$ represented as a set of features $A = {a_1, a_2,\dots, a_n}$, and the class $Y_m$ having a binary value representing either a \textit{``late''} or \textit{``on time''} state for $I_m$.
\end{definition}

In order to prepare the dataset for the model training, we defined class $Y_m$ for each invoice $I_m$. The definition of class $Y_m$ as \textit{``on time''} or \textit{``late''} is done as follows: 

\begin{equation}
\small
  Y_m=
    \begin{cases}
      \text{on time}, & \text{if payment at most 5 days from due date} \\
      \text{late}, & \text{otherwise}
    \end{cases}
\end{equation}

Thus, an invoice is considered overdue if payment occurs more than 5 days after the due date. The main reason for considering this time window is the time required to process payments in the client system. This interval was elicited during one of the meetings we had with client's subject matter experts (SMEs) to understand the problem, processes, and work flow of the collection activity.

\section{Data Source and ETL} \label{sec:data}

The dataset received from the client has 91,562 invoices from 6 countries from Latin America plus the United States, with 2,229 customers over dates ranging from November 2018 to November 2019. The distribution of invoices by country is presented in Table~\ref{tab:invoicecountry}. Since we have some countries with low representativeness, after analyzing the payment behaviour of the countries with high representativeness in the dataset, we decided to develop only one model for all countries, instead of one model for each country.

\begin{table}
\centering
\begin{tabular}{ lrr } 
\toprule
Country & \# invoices & \# customers \\ 
\midrule
United States & 26,506 & 4,276 \\
Argentina & 220 & 27 \\ 
Brazil & 17,510 & 785 \\ 
Chile & 11,634 & 324 \\ 
Colombia & 16,302 & 514 \\ 
Ecuador & 6 & 1 \\ 
Mexico & 19,484 & 578 \\ 
\bottomrule
\end{tabular}
    \caption{Data distribution for Latin America and North America shown by country. Some countries are very under-represented in the data, thus we aimed at building one model for all countries instead of a particular model for each country.}
    \label{tab:invoicecountry}
\end{table}

The dataset only contains information about payments, specifically about invoices. For instance, invoice value, country code, customer number, etc. One of important challenge of using this dataset is that it has no relevant information about customers. Therefore, information such as industry sector and balance sheets are missing from the dataset. The dataset only has a unique identifier used to differentiate customers.
Since the scope and time of the project does not include gathering information about customers, and due to privacy constraints, we decided to proceed with this anonymized data, and work only with invoice information.






One of the biggest challenges in projects such as the one described here, is how to transform the invoice data to enrich it with relevant features in order to build a machine learning model.
Being the result of manual data processes, and as is usual with real datasets, we found a large number of missing values and incorrect data in several fields that would be important for machine learning models. Additionally, the dataset was compiled from legacy systems of branches from all over the world, and the result of a data consolidation effort by the client.  
In order to extract the most valuable features from the data, we performed several discussions with client's SMEs and collectors, looking at previous literature on AR, and performed a thorough, detailed examination of the dataset.

We started with traditional invoice-level features, such as invoice amount, issue date, due date, settled date, etc. 
In order to enrich our model with more significant information, we performed feature extraction to build the late invoice payment model.

We used historical data to create aggregate features that could bring more meaning to our set of invoice-level features, as the use of aggregate features increase significantly the amount of information about payment~\cite{zeng2008using}.
However, some of the features recommended in the literature~\cite{zeng2008using} did not work in our case, specially due to some data related issues. One such example is the computation of ratios that, due to the amount of missing information in some fields, increased the number of invalid (such as null or \textsc{NaN}) values in the data. Additionally, some features, such as a category that specified whether an invoice was under dispute or not, had consistency issues, due to being manually entered information. 

On the other hand, we incorporated some features related to the recent payments in order to capture customer behavior. Based on our careful analysis of the data, we observed that recent payments influence more in the payment behaviour than older payments. In other words, the recent payment behavior of a customer has more predictive power than the complete payment history.
In this particular case, we noticed that encoding whether a customer had paid each of their last three invoices had more predictive power. In addition to that, we computed as features the percentage of paid invoices, payment frequency, number of contracts related to each invoice, and standard deviation of late and outstanding invoices. The list bellow shows the constructed features as well as the respective descriptions.

\begin{itemize}
    \item \textit{paid invoice}:  Value indicating whether the last invoice was paid or not; where 1 means paid, 0 means not paid, and -1 indicates null value (possibly due to this customer being a first time customer).
    \item \textit{total paid invoices}:  Number of paid invoices prior to the creation date of a new invoice of a customer. 
    \item \textit{sum amount paid invoices}: The sum of the base amount from all the paid invoices prior to a new invoice for a customer.
    \item \textit{total invoices late}: Number of invoices which were paid late prior to the creation date of a new invoice of a customer.
    \item \textit{sum amount late invoices}: The sum of the base amount from all the paid invoices which were late prior to a new invoice for a customer. 
    \item \textit{total outstanding invoices}: Number of the outstanding invoices prior to the creation date of a new invoice of a customer. 
    \item \textit{total outstanding late}: Number of the outstanding invoices which were late prior to the creation date of a new invoice of a customer. 
    \item \textit{sum total outstanding}: The sum of the base amount from all the outstanding invoices prior to a new invoice for a customer. 
    \item \textit{sum late outstanding}: The sum of the base amount from all the outstanding invoices which were late prior to a new invoice for a customer. 
    \item \textit{average days late}: Average days late of all paid invoices that were late prior to a new invoice for a customer. 
    \item \textit{average days outstanding late}: Average days late of all outstanding invoices that were late prior to a new invoice for a customer 
    \item \textit{standard deviation invoices late}: Standard deviation of all invoices that were paid late. 
    \item \textit{standard deviation invoices outstanding late}: Standard deviation of days late of all outstanding invoices that were late prior to a new invoice for a customer.
    \item \textit{payment frequency difference}: Amount of times the customer did a payment. Intention here is to identify customers that payed more invoices (payment could be 30, 45, 60 days). 
\end{itemize}

The next step in our ETL process was to handle missing values. We had to do this for invoices that did not have too much historical information according to our features.
One such case was that, if we had a null value for the total sum of invoices, we just replaced it with zero. However, in a few cases it was necessary to take into consideration what the feature meant and its representation. For average days late, for example, we could not fill with zeros, as it would be an indicative of good payment behavior. In such cases, we used the mean value of the feature as a replacement rule for missing ones. 

With the dataset cleaned up and properly set up, we could start to work in the best way to split the data into train and test sets. 
To prevent data leakage, that is to create an accurate model to make predictions on new data, unseen during training, we split our dataset considering time. Thus, the split into train and test sets was based on time of invoice creation. For training, we considered data ranging from November 2018 to April 2019 and test from May 2019 to November 2019. Table~\ref{tab:modelrange} presents in details the amount of invoices that we had in each sample of our dataset. As we can see, the distribution of invoices between late and on time is a somewhat imbalanced towards late invoices, except that in the test set we have the same distribution between late and on-time invoices.  

Once we had the training set, we used cross-validation to tune hyper-parameters. The folds were separated using a time series split, based on invoice creation date. This is needed because we want the distribution of data in the validation set to be as close as possible to the data in the test set, so that performance measures in the validation set is predictive of performance in the test set. Therefore, a random shuffle would be a bad option in this case, since it would mix more recent invoices with older ones in both datasets, giving an advantage to the model during training and validation which would not bbe reflected with new, unseen data.

\begin{table*}[htb]
\large
\begin{center}
\begin{tabular}{ lrrrrc } 
\toprule
 Dataset & invoices & Late & On Time & Baseline & Period \\ 
\midrule
Train & 71,545 & 61.17\% & 38.83\% & 61.17\% & 2018-11 - 2019-06 \\ 
Test & 19,630 &  54.51\% & 45.49\% & 54.51\% & 2019-07 - 2019-11 \\ 
\bottomrule
\end{tabular}
\end{center}
    \caption{Data distribution for training and test sets. The classes are balanced and our baseline is close to 61\%. We split the data using time, since we cannot use future data to make predictions. We use 70\% of available data to train the model and the other 30\% we split in test and validation data. We did a time series split since the temporal characteristic of the that does not allow traditional cross-validation method. }
    \label{tab:modelrange}
\end{table*}

\section{Modeling Approaches and Results} \label{sec:experiments}

As explained before, our problem was defined as a binary classification problem to predict whether an invoice will be paid on time or late. Although we stated the problem as predicting classes, a wide range of models return probabilities instead of just labels. This is crucial in order to do a prioritization list and rank customers with higher chances of default. Also, since the model is planned to be deployed in a client service that ultimately will need to retrain and update the model, it is important that we use a powerful model in terms of scalability, while being able to handle missing values, and with results that are easy to understand while being easy to retrain, so that non-machine learning experts could have a sense about what is going on with the data and the model.

Most of features came from historical data, for example, \textit{sum amount late invoices}, \textit{total invoices late} and so on. In order to calculate these features for an invoice, we needed to define a maximum period of time that the system would consider to look back. This period is different from our trained dataset that defines which invoices we will consider. To define the best range of time to look back to calculate the features, we created a parameter that we call \texttt{window size}, represented by the letter $w$.
In short, window size will be the number of months prior to an invoice that we will consider to calculate our features values.
        
One might wonder why limit past data to a window, instead of using the complete, historical data. 
The problem is that, as time passes, the statistical distribution of the features also changes and reduces the accuracy of the model, since all the models considered work on the assumption that data is independent, and identically-distributed. In the machine learning and predictive analytics realm this is known as concept drift. Therefore, it is necessary to work with boundaries and to focus on getting information from the most recent past, representing the most recent customer behavior. 

In order to create a more robust model and to make sure that we are using the correct time range, we created ten datasets with $w$ ranging from 3 to 12 months to perform tests and see how many months do we need to consider. We decide to use as $w$ values from 3 months or above, because in our feature engineering we have a feature that looks at the three last payments that are mostly in the 3 months range and we know that our data is very susceptible to data drift and is susceptible to external changes, such as changes in the economy, politics, and policies.

We tested our data with six different classification methods: Naive Bayes, Logistic Regression, k-Nearest Neighbors ($k$-NN), Random Forest~\cite{Breiman:2001}, Gradient Boosted Decision Trees (XGBoost)~\cite{FRIEDMAN2002367}, Deep Neural Network based on the fastai library (DNN)~\cite{Howard_2020}.
For each classifier, we performed cross-validation for hyper-parameter selection. Below, we describe how each model was built and the set of hyper-parameters used for model selection.

\paragraph{Naive Bayes}

We used a Gaussian Naive Bayes implementation of the scikit-learn 0.22 library. Due to the simplicity of the implementation, we did not build a grid of hyper-parameters to search through for this algorithm.

\paragraph{Logistic Regression}

We used the Logistic Regression implementation of the scikit-learn 0.22 library. Hyper-parameters for this model were (with parameters in bold representing the best values found by a cross-validated grid search): penalty chosen from $[L1, \mathbf{L2}]$, C (inverse of regularization strength) chosen from $[\mathbf{0.5}, 1, 5, 10, 20, 50]$, class weights chosen from $[\text{balanced}, \textbf{None}]$, and maximum number of iterations chosen from $[10, 50, \mathbf{100}]$.

\paragraph{k-Nearest Neighbors}

We used the k-Nearest Neighbors implementation of the scikit-learn 0.22 library. For hyper-parameter search, we chose the number of neighbors from elements of the set $\{n : 29 \le n < 100\}$, and the best number of neighbors found set to 49.

\paragraph{Gradient Boosted Decision Trees}

We used the XGBoost library and performed a grid search for model selection. Hyper-parameters were, with the parameters in bold representing the best found values for each hyper parameter: learning rate was set to $\mathbf{0.01}$, data subsampling in the set $[0.7, \mathbf{1.0}]$, column sampling by tree and by level in the tree in the set $[\mathbf{0.7}, 1.0]$, L1 regularization in the set $[\mathbf{1}, 10]$, the number of estimators was set to $\mathbf{100}$, and the maximum tree depth was the set $[3, 7, \mathbf{15}]$.

\paragraph{Deep Neural Network}

To define a deep neural network for classification, we used the fastai library version 1.0.61, which makes use of the PyTorch library, for which we used version 1.2.0. While in the other models we used a one-hot encoding for differentiating countries, with the neural network we used an embedding layer to model countries. After processing the countries through the embedding layers, we apply a dropout layer to the embedding outputs prior to concatenating its outputs with the other input data and process the data through a regular multi-layer perceptron (MLP). Each layer of the MLP is composed of blocks of Batch Normalization~\cite{ioffe2015batch}, Dropout~\cite{srivastava2014dropout}, Linear, and ReLU activation, with the first block skipping the Batch Normalization and Dropout, and the last block skipping the ReLU activation.

To find an architecture and hyper parameters, we performed a cross-validated search. We searched through networks with 2, 3, and 4 layers in the MLP part, with the best network found being the one with 4 layers. On top of that, we performed tuning of the class weights of the data, to mitigate the imbalances of the data. The weight of class 0 (on-time payment) was kept at 1, while the weight of class 1 (delayed payment) was searched through the set $[1.6, 2, 3, 4]$, with the best value found set to $3$. 

\subsection{Results}
We observed that XGBoost, Random Forest and DNN achieved the best accuracy with $w=4$, while k-NN and Logistic Regression with $w=10$. Naive Bayes achieved its best at $w=3$. This implies that, using more data than we really need to impacts model performance negatively, degrade it, instead of making it better. 

Indeed, as we can see, a small number of months works better than a large one. This is specially due to the concept drift in the data. In our dataset, customers are performing better over the months, that is, they are paying invoices less late than in the starting point of the dataset. Thus, using all the data to calculate historical features would insert an incorrect bias in the model.

\begin{figure}
    \centering
    \includegraphics[width=0.48\textwidth]{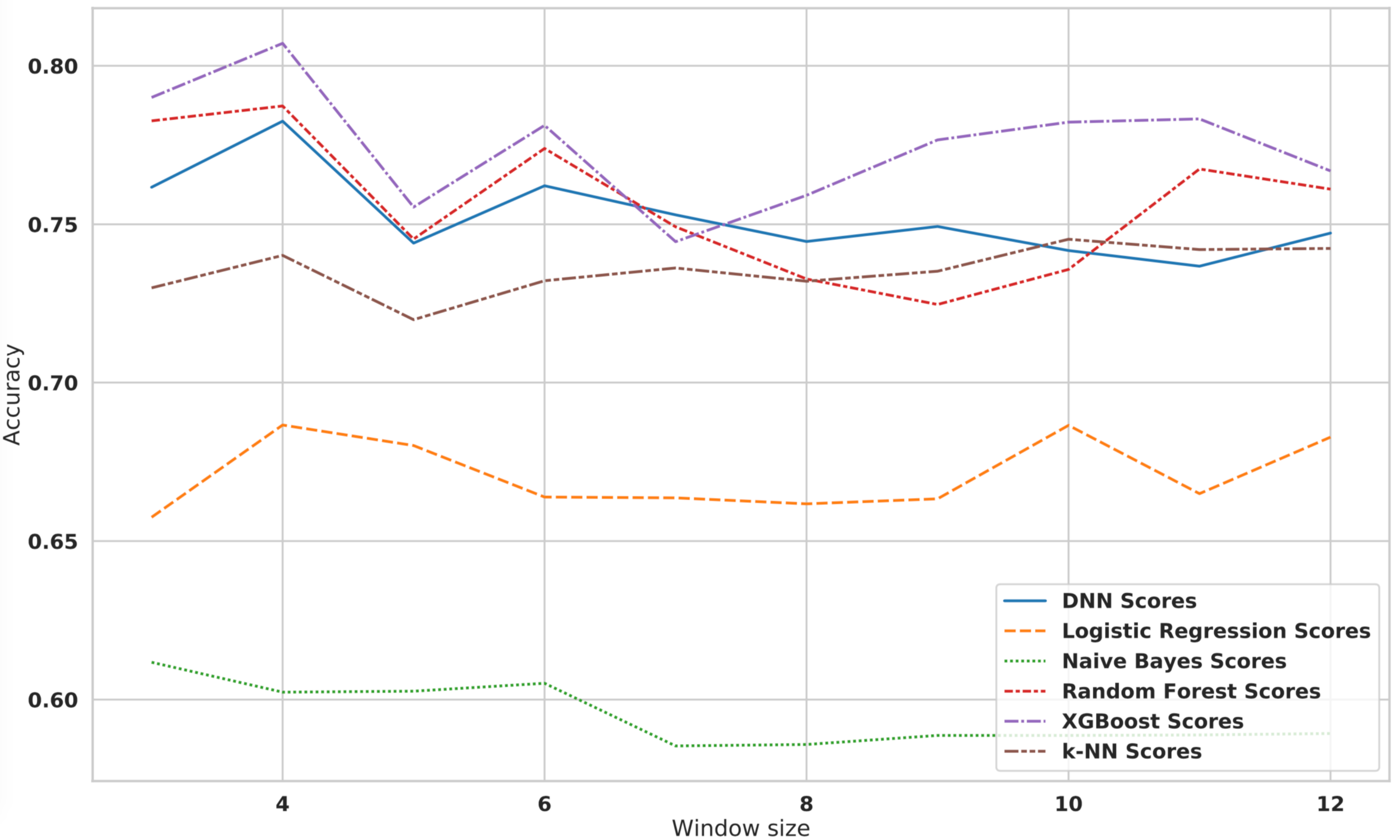}
    \caption{Plot showing accuracy for all tested models with all the $w$ generated. The ensemble methods are the ones with high accuracy in all $w$, being $w=4$ the best result. Naive Bayes and Logistic Regression are the ones that scores worst. We see that more we accumulate our historical features, the worst the models perform.}
    \label{fig:modelacuracy}
\end{figure}

Figure~\ref{fig:modelacuracy} shows the results for all models and $w$ values. Compared to the baseline score (54.51\%), we can see that the models perform significantly better, specially the ensemble learning models (Random Forest and Gradient Boosting) are the ones with higher accuracy. 

\begin{figure}
    \includegraphics[width=0.48\textwidth]{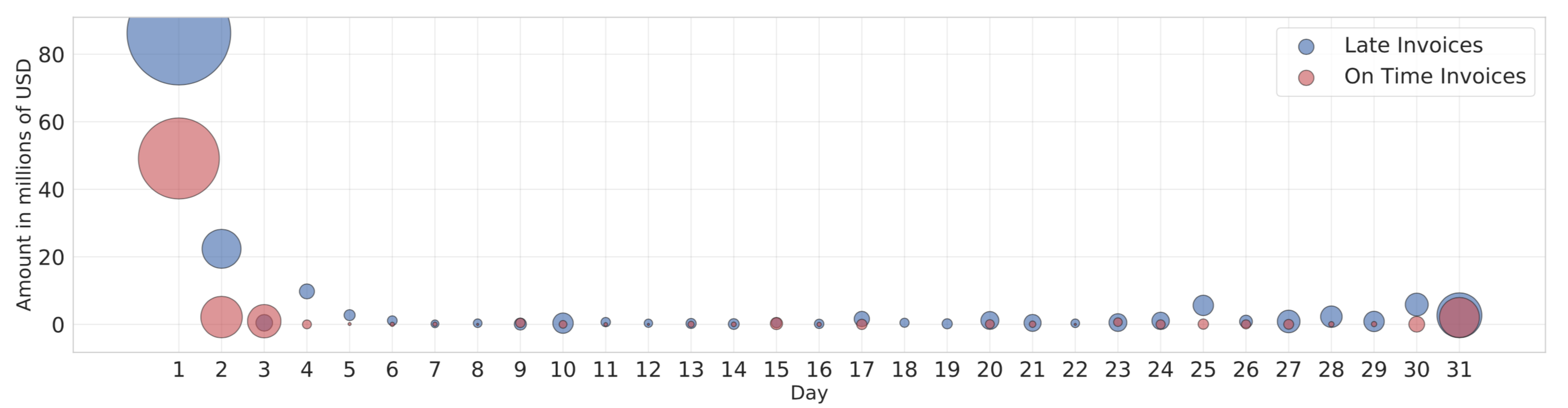}
    \caption{Invoices to be received over one month, distributed over the payment data. Each point could have one or more invoices.}
    \label{fig:montlycollectpredict} 
\end{figure}

We also test the accuracy per month with XGBoost with $w=4$, since the model, in practice, will be used monthly to compute the invoice prediction over monthly. Figure \ref{fig:montlycollectpredict} present this result. We can see that for the two last month the accuracy was up to 85\%. 

\begin{figure}
    \centering
    \includegraphics[width=0.48\textwidth]{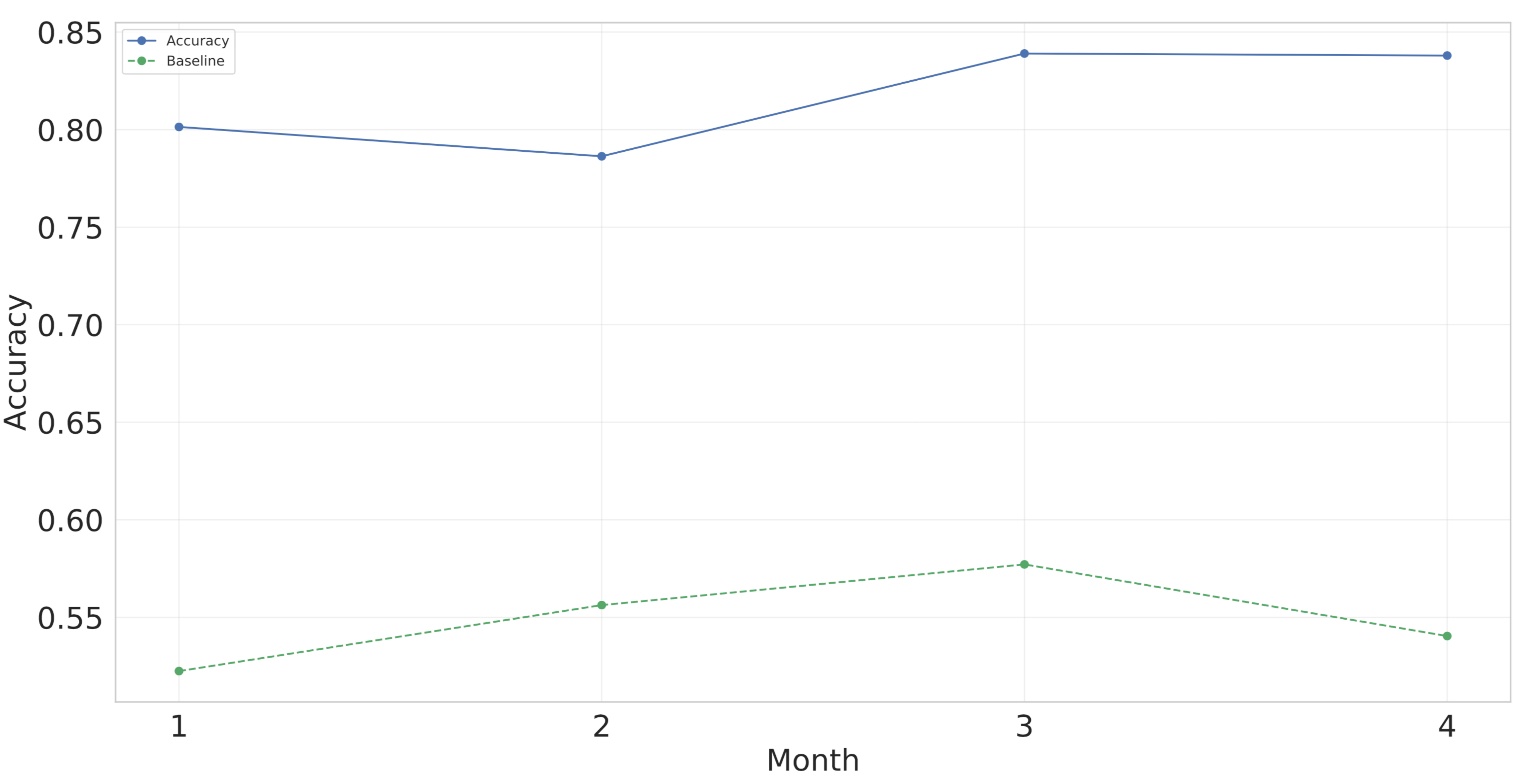}
    \caption{Accuracy per month in the test set with XGBoost and $w=4$. }
    \label{fig:acuraccypermonth}
\end{figure}

Given that our model returns a probability score rather than a label indicating whether it is from class late or on time, we can rank the targets and use it to plot Receiver Operating Characteristic (ROC) curves in order to compare models. They depict the performance of a classifier comparing the True Positive Rate vs False Positive Rate. True Positive Rate, also known as recall metric, indicates that given a true class, for example, class late, what is the percentage of samples classified as late compared to the ground-truth, that is, the total number of instances from class late. On the other side, False Positive Rate is the percentage of falsely reported positives out of the ground-truth negatives (class on time). Intuitively, the ROC curves will give a guidance to understand how well a model is performing based on a ranking. That is, if we have a high lift point on the curve, it demonstrates that invoices with late labels have higher probability score of being late (as expected).

\begin{figure}
    \includegraphics[width=0.45\textwidth]{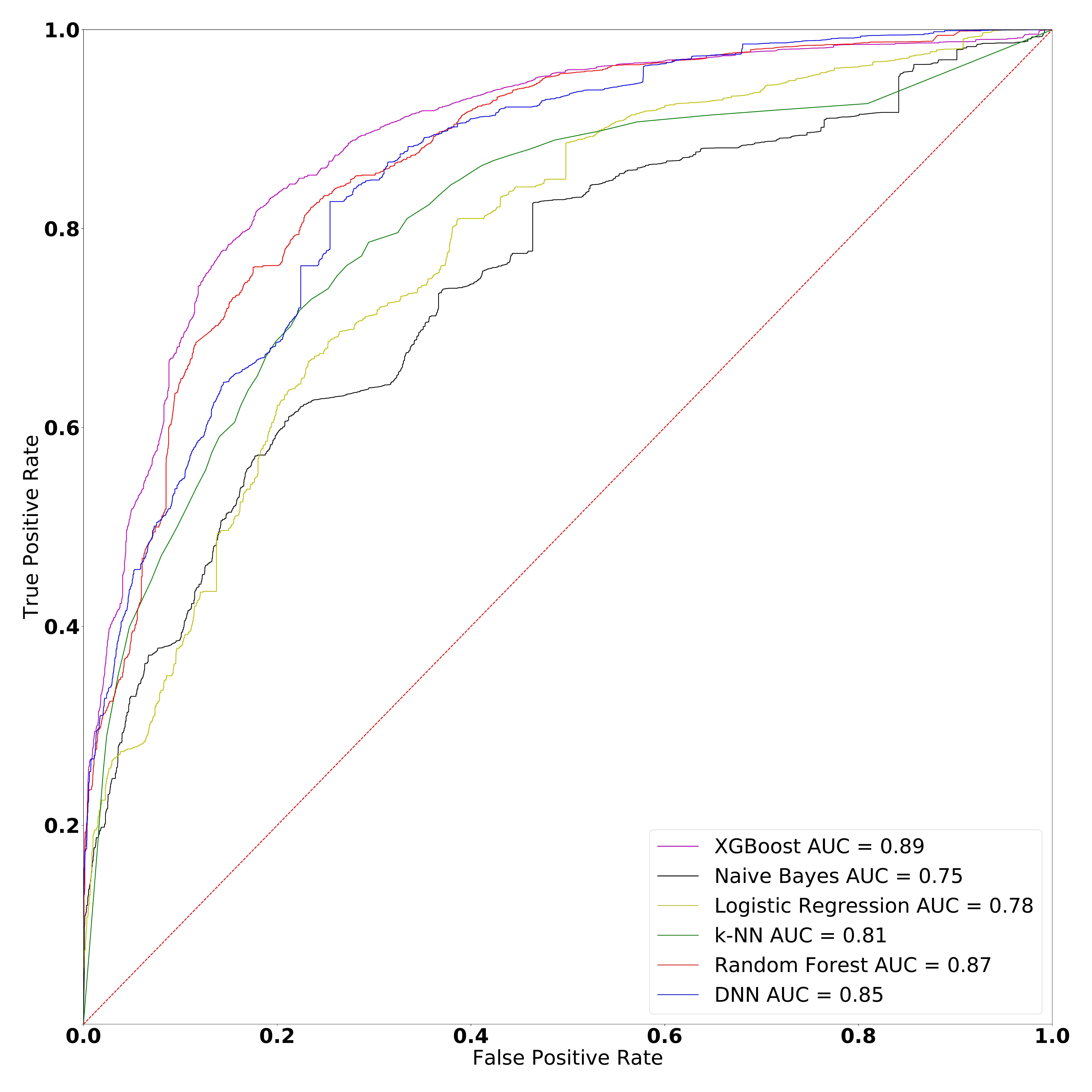}
    \caption{ROC Curve of all five methods we tested in our test dataset with $w=4$.}
    \label{fig:rocv}
\end{figure}

In order to measure not only graphically but also quantitatively, we used the area under the curve metric (AUC) as well. AUC is a metric that calculates the area under our ROC curve, i.e., it as a way of calculating the lift point explained above. Figure~\ref{fig:rocv} shows the ROC curves for each model and the corresponding AUC score for test and validation.

Clearly, the best models are the Random Forest and Gradient Boosting. Therefore, we developed the predictive invoices label system based on an ensemble approach using both models.

With the model trained we finally can use if to help collector optimize AR processes. One way to do that is Figure \ref{fig:montlycollectpredict}, which could be a interactive visualization showing the invoice distribution due date over the month. Each day we have the invoices that are due for data day label as late or on time. Each bubble could be link to a list of invoices (and clients) where the ones marked as late the collector could take a action. For instance, for the late invoices the collector could give a call remembering the payment.  For each month when a new invoice is created the model will give a prediction if the invoice will be payed Later or On Time, with this in hands, collectors can have a better vision of the process and focus in clients that are predict to be late and only follow the invoices that are predict to be on time to check if they were really payed.  As we will show in the Section \ref{sec:ranking} we also proposed a new way to prioritizing the clients that before the use of machine learning were sorted mainly by the amount of money that they are in debt. Now, with the model we can use the probability of an invoice be payed late to adjust this sort. 

\subsection{Model Robustness}

Although the entirety of the dataset was created by same bank, this doesn't mean the data is homogeneous. The bank has customers from different countries and each region uses different legacy systems, especially when we compare North America with Latin America. Therefore, it's possible to have different behaviors from each country due to local economic factors and operational differences. These differences could have an impact in the model, showing the importance to have a robust model that can learn this distinct patterns enabling the creation of a general model, capable of working with the complete dataset and, consequently, with all the regions present in the dataset.

\begin{table}
    \centering
    \begin{tabular}{lccc}
        \toprule
         Dataset & NA Accuracy & LA Accuracy & General Accuracy  \\
         \midrule
         NA + LA & 82.02 & 79.87 & 80.71 \\
         NA & 72.27 & - & 72.27 \\
         LA & - & 78.91 & 78.91 \\
         \bottomrule
    \end{tabular}
    \caption{Model Accuracy using data from NA and LA together and separately. In the first line we show a  model that was trained with LA and NA data together and tested with each region individually (first two columns) and both together (third column). NA (the second line) represents a model trained only with data in from the NA, and tested only with NA data. LA (third line) is the model trained only with data from LA and tested only data from LA. We can see that the accuracy is better using data from both LA and NA together.}
    \label{tab:modelNALA}
\end{table}

In Table \ref{tab:modelNALA} we present an experiment with the XGBoost model and $w=4$ to test the robustness of the model trained with the complete data contrasted with using only NA or LA data. The NA+LA line shows a model that was trained with the full dataset, and shows the evaluation performance of all the test set, and also for only the NA region, and only for the LA region. We can see that NA has better accuracy on test, 82\%. The other two models were trained exclusively with NA and LA and tested only with data from their respective regions. We can see that the accuracy is lower using separate models both in LA and NA. Using both regions together we were able to improve the accuracy and make the model more robust. This also represents a benefit to the client, since the client is planning on expanding to new geographies, and a single model is easier to maintain than having several, separate ones. 






\section{Invoice Prioritization} \label{sec:ranking}

In the previous section we demonstrated that we can effectively predict the probability of an invoice being late. Identifying invoices that are likely to be delinquent at the time of creation enables us to steer the collections process, thus helping to save resources~\cite{zeng2008using}. However, our objective is not to make decisions based on invoices. Rather, we intend to make decisions based on the clients that hold the invoices. Furthermore, considering that resources are finite, we can not work with the hypothesis that all invoices have the same impact. In other words, we have to associate the probability of an invoice being late with its value in dollars. This way, we will be able to measure the delinquency risk based not only on probabilities, but also based on total invoice amount.

Currently, collectors can be considered agents that follow a greedy policy based on the value of invoices. In other words, collectors will prioritize invoices based on the total amount of dollars that is overdue. 
As an example, consider the following hypothetical situation: suppose there is an invoice $I_1$ with low probability of being late, such as $P_{I_1}= 0.2506$. But $I_1$ has a high value, such as $V_{I_1}=\$1,000,000.00$. Suppose further that there exists an invoice $I_2$ that has a high probability of being late, $P_{I_2} = 0.9358$, and a lower value, such as $V_{I_2}=\$300,000.00$. In this situation, without our model, $I_1$ would have higher priority than $I_2$.


As we show in Equation~\eqref{eq:rankingi}, we propose to take into account an invoice's risk of being late, multiplying its probability of being late by its value. In this way, we can continue prioritizing big customers and, at the same time, we save efforts on customers which will probably pay the invoice on time. 

\begin{equation}
\mathcal{R}_{I_i} = V_{I_i}*P_{I_i}( Y = Late )
\label{eq:rankingi}
\end{equation}

Next, we need to associate the invoice's level of information with a customer's level of information. We assume that clients will be contacted based on their total amount of risk, rather than based solely on a single invoice. To create a customer ranking, we decided to average the risk of invoices by customer as shown in Equation~\eqref{eq:rankingc}.

\begin{equation}
\mathcal{R}_{C_j} = \frac{1}{N} \sum\limits_{i=1}^N \mathcal{R}_{I_i}
\label{eq:rankingc}
\end{equation}

In order to compare our new prioritization ranking with the previous, greedy policy followed by collectors, we use Kendall's $\tau$~\cite{kendall1948rank} as a metric to compare the number of pairwise disagreements between two orders. Values close to 1 indicate strong agreement, while values close to -1 indicates strong disagreement. Our new ranking has $\tau = -0.003$ which means that we change about $50\%$ of the ranking order. 

To investigate whether the ranking function proposed in this section can be effective in increasing recovery, we simulated how the system would behave when customers were contacted using the greedy approach, and using the proposed ranking functions. To set up the simulation, we assumed collectors can contact a finite amount of customers per month. We performed the simulation considering $n = {100,200,300}$ calls. Since we don't know the real impact a contact has on customers, we assigned a probability $p$ of the contact being successful and resulting in the payment of an invoice. We varied $p$ from $0$ to $1$ using a step size of $0.1$. The probability of $0$ serves as a sanity check for the simulation, since it means that calls have no effect and, therefore, both rankings should result in the same amount of money collected. Although it is unrealistic that \emph{all} customers will have the same probability of reacting positively to a contact, this is still a useful model to understand the overall impact of the new ranking function.

\begin{figure*}
    \centering
        \includegraphics[width=0.9\textwidth]{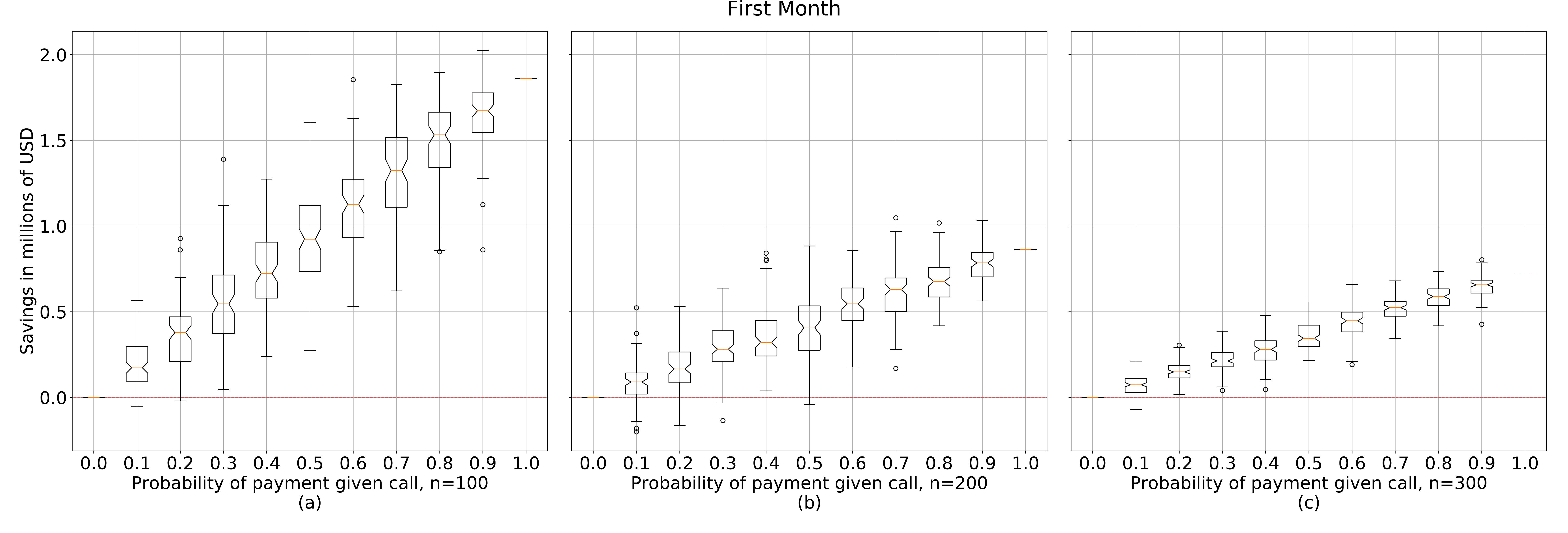} \\
        \includegraphics[width=0.9\textwidth]{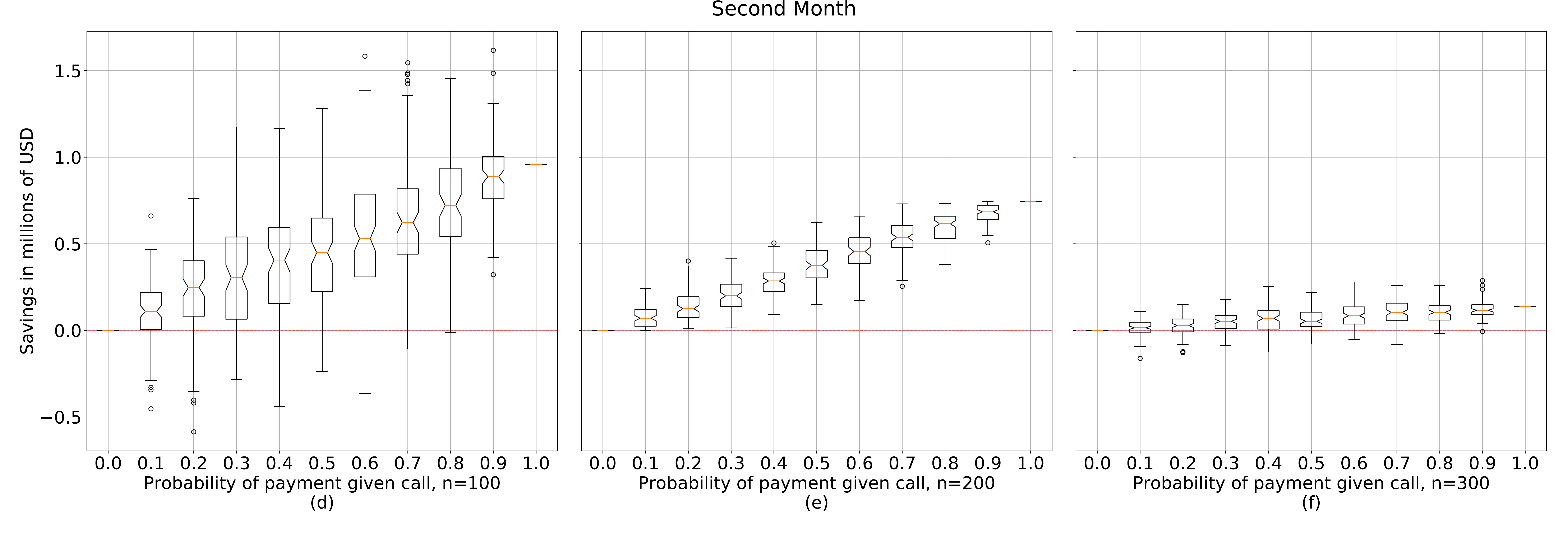} \\
        \includegraphics[width=0.9\textwidth]{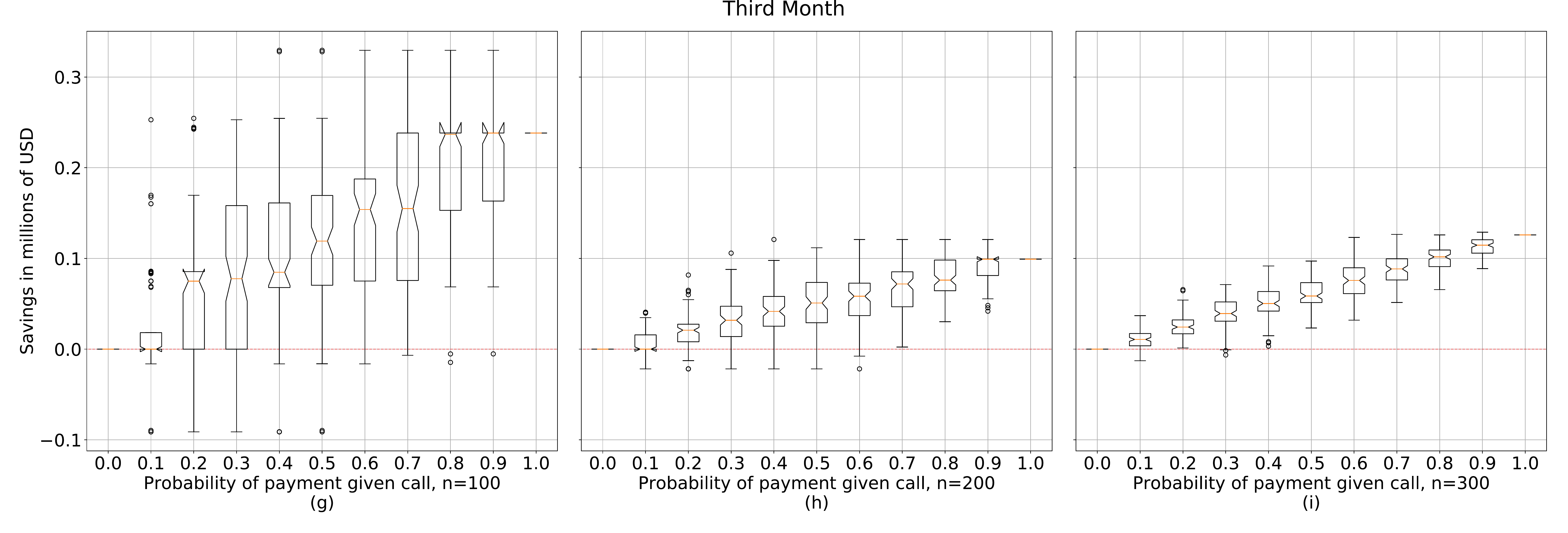} 
    \caption{Simulation of the probability of a invoice being payed given a call from collector. The $n$ represents the number of clients that a collector will call $100, 200, 300$. We also vary the probability of payment given a call on axis $x$ and the savings in millions of USD in axis $y$ compared with traditional ranking.}
    \label{fig:simulation}
\end{figure*}

Figure~\ref{fig:simulation} shows the results of our simulation. In the figure, we show the \emph{difference} in savings (money collected) when using the proposed ranking compared to the greedy ranking function. Each box plot in the figure aggregates the data of a hundred independent simulation runs. We performed the simulations for three different months in our dataset. Each row of Figure~\ref{fig:simulation} shows data for a particular month, while each column shows the impact for a different $n$, described above. The results show that, in every scenario, the ranking function created with our model results in a better ranking, which increases the money collected up to approximately 1.75 million dollars. Smaller values of $n$ show higher savings, which makes sense: since collectors can't contact many clients, the impact of choosing \emph{which} clients to contact becomes more important. This is in line with reality, since collectors have a limitation in the number of clients that they are able to call in a month. 

As can be seen in the figure, the bigger $p$ gets, the bigger are the savings. This happens because, as the probability increases, the impact of correct prioritization becomes more apparent, since a better ranking might be able to change invoices that would become overdue into invoices that become paid on time. Impact is higher on the first month because we have more data for that month. Our dataset comprises 7089, 4260 and 2626 entries for months 1, 2 and 3 respectively. 


\input{conclusion}
\bibliographystyle{ACM-Reference-Format}
\bibliography{igfbibliografy}

%


\end{document}

%% file: introduction.tex
\section{Introduction}

%

The invoice-to-cash process involves various steps, from invoice creation to customer's debt (payment) settlement or reconciliation. One key step of this process is the collection of accounts receivables. Accounts receivables (AR) refers to the invoices issued by a company for products or services already delivered but not yet paid for by its customers. Properly managing AR is a core accounting activity and concern of any company, pertaining to its cash-flow. 

Despite the widespread use and adoption of information technologies in recent years across domain and industry applications, particularly machine learning techniques, there are still companies that manage internal processes in the same ways they did in the past: with paper and pencil. 


In this work, we present a case study carried out in partnership with a multinational bank (hereafter also referred to as client).
In this case study, we sought for innovative ways to proactively identify overdue ARs with high probability of being paid such that its managers and executives could take appropriate actions (such as, reaching out to those customers and collecting those ARs). As an international bank, it operates in multiple countries; however, this project focuses exclusively on its customers based in Latin America and the United States. 

The collection activity is performed by analysts, also known as collectors. Collectors are responsible for charging the bank's customers (hereafter referred to as customers) and, in turn, improving these customers' experience relative to the payment processes. In the bank, each collector deals with approximately a hundred customers and these customers are allocated according to the seniority level of the collector, meaning that senior collectors are responsible for bigger accounts and contracts. In collecting ARs (i.e. the activity of charging customers), collectors receive daily a list prioritizing customers to contact. This list takes into account a customer's debt based on all of its overdue invoices. Nonetheless, it ignores the customer's payment behavior, such as, whether this customer regularly pays on time or not. Usually, customers are contacted at a fixed schedule before due dates, irrespective of whether a particular customer regularly pays its invoices on time or not. Neither does it differentiate a recurrent from a sporadic payment behavior, such as an occasional financial problem faced by a customer. However, in the end, all of this detailed information about customers' payment behaviors lays in individual collectors' minds, being utilized just in an \emph{ad-hoc} manner. 

\begin{figure}[htb]
    \centering
    \includegraphics[width=0.45\textwidth]{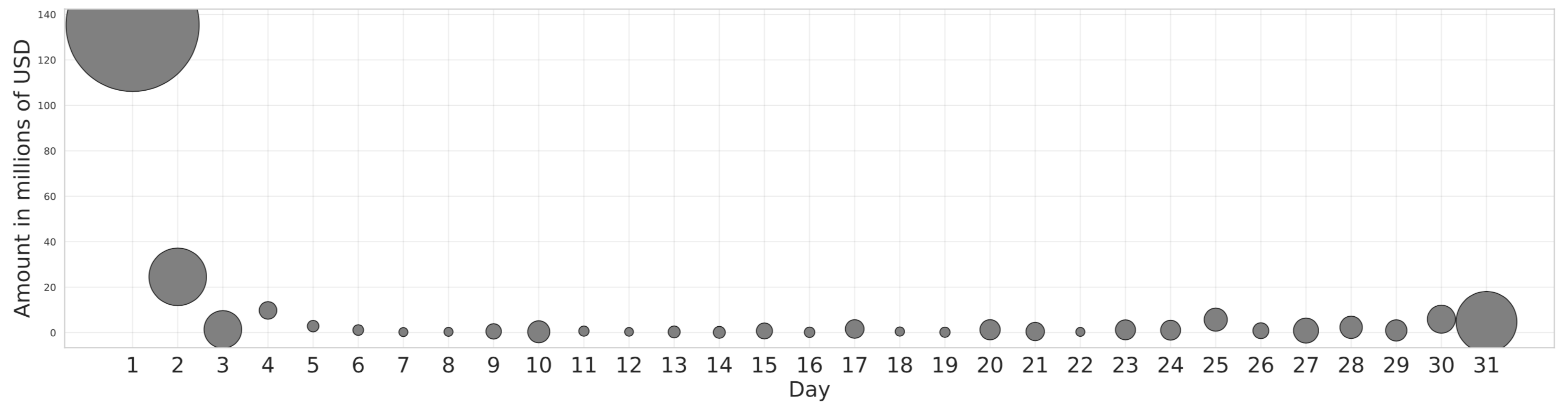}
    \caption{Receivables over one month, distributed over the payment data. Each point can have one or more invoices, with bigger points representing a bigger number of invoices.}
    \label{fig:montlycollect} 
\end{figure}

Figure \ref{fig:montlycollect} represents what collectors face every day, which is dealing with a large number of invoices from several clients in a month. Each bubble represents a set of invoices with the due date for that day, the size of each bubble represents how many invoices are there in any particular day, potentially from different clients. The position of the bubbles are the amount of money to be receive in a given day. Looking only at that Figure, it might become hard to prioritize which clients should be contacted first. Such a representation leaves information out, such as which clients are more likely to pay late. Therefore, the easy route might be to go after the larger amounts of money first, since collector performance is measured by the amount of money they recover. A larger sum of money does not necessarily represent higher risk clients, though. Hence, predicting invoice which invoices are most likely to be paid next can be a solution to better allocate resources, impacting positively cash flow estimation, essential to achieving financial stability.


In our research, we assert that by providing insights as to how to prioritize contacting clients based on the probability of late payment helps collectors make more effective and efficient decisions. The objective is to help them make more assertive and timely decisions by means of focusing their collection actions on invoices that would have a greater financial return, while at the same time focusing on those most likely to make a payment. To this end, the system should provide a personalized, ranked list of customers whom to contact, taking into account a collector's own list of customers and their payment behavior.



Predictive modeling approaches are widely used in a number of related domains, such as credit management and tax collection~\cite{abe2010optimizing}.
The problem of predicting invoice payment has been traditionally tackled using statistical survival analysis methods, such as the proportional hazards method~\cite{Lee:2013}. 
Survival analysis is a statistical method for analyzing the expected duration of time until one or more events happen, such as death in biological organisms and failure in mechanical systems.

Dirick et al.~\shortcite{Dirick2017} tested several survival analysis techniques in credit data from Belgian and Great Britain financial institutions. Survival analysis techniques were also used to model consumer credit risk~\cite{cao2009modelling,cheong2018customer,rychnovsky2018survival}. The aforementioned pieces of work focus on predicting \emph{when} an event may occur, rather than \emph{whether} it may occur or not. This aligns with our interest in analyzing time to an event; thus, a survival analysis approach is a reasonable technique for tackling the problem at hand. 

Smirnov~\shortcite{smirnov2016modelling} concluded that \textit{Random Survival Forests} models, which additionally uses historical payment behavior of debtors, perform better in ranking payment times of late invoices than traditional \textit{Cox Proportional Hazards} models. Although the \textit{proportional hazards} model is the most frequently used model for survival analysis, it still has a number of drawbacks, such as having a baseline hazard function which is uniform and proportional across the entire population, as explained by Baesens et al.~\shortcite{baesens2005neural}.

Invoice payment prediction could also be modeled as a classification problem, but there is just a small body of work that addresses this problem. One of the few works that investigate this is the one by Zeng et al.~\cite{zeng2008using}, where the authors formulate the problem as a traditional supervised classification and apply existing classifiers to it. They divided the clients into four different classes related to payment delays: on time, 1-30 days, 31-60 days, and +60 days. These classes are usually related to AR processes and counter measures for addressing late invoices. Similarly, Bailey et al.~\cite{bailey1999} analyze several strategies for prioritizing collection calls and propose to use predictive modeling based on binary logistic regression and discriminative analysis to determine which customers to hand over to an outside collections agency for further collection processing.


Tater et al.~\shortcite{tater2018prediction} propose a different approach to the problem and instead of predicting invoice in accounts receivable, they focus on accounts payable, working on invoices that were already delayed. Similarly, Younes~\shortcite{younes2013framework} focuses on accounts payable case and attempts to address the problem of invoice processing time, understanding the overdue invoices and the impact of delays in the invoice processing.
Abe et al.~\shortcite{abe2010optimizing}, in addition, propose a new approach for optimally managing the tax and, more generally, debt collections processes at financial institutions.


The prototype herein described aims at devising and developing a tool employing state-of-the-art Artificial Intelligence (AI) algorithms and techniques for creating a ranked list of (potential) overdue accounts for each collector, based on different criteria, such as the highest probability of payment in the short-term, payment behavior patterns, and the like. 
It thus aims at optimizing collectors' actions and thus improving the payment rate of these accounts.

The key contributions of this paper are: 
\begin{itemize}
\item The use of machine learning to predict with high accuracy the status of invoices (late or on time), allowing the bank to better estimate how much money will be delayed in cash and work pro-actively to avoid late payments; 
\item The use of historical and temporal features to improve accuracy of models and an extensive comparison of Machine Learning models applied to this problem;
\item An effective new way for ranking customers to be prioritized by collectors not only by taking into account the volume of money, but also the probability of being late; 
\item Our simulations show that adoption of our model to prioritize the work of collectors saves up to $\approx$1.75 million dollars per month.
\end{itemize}

The paper is organized as follows. Section \ref{sec:problem} defines the problem formally. Section \ref{sec:data} characterizes the data set used in the work and the ETL (Extract, Transform, Load) process. 
Section \ref{sec:experiments} shows the modeling approach applied in the problem and the results obtained so far. Section \ref{sec:ranking} presents the prioritizing list proposed to client and the simulations showing that the model and the new ranking are able to saving a large amount of money.  
Finally, Section \ref{sec:conclusion} discusses outcomes and concludes our work. 

%% file: conclusion.tex
\section{Conclusion} \label{sec:conclusion}

In this paper, we built a model that computes the probability score of an invoice being overdue in the context of AR practices. This is critical when dealing with a very large set of invoices, which in turn requires collectors to rank customers and focus on those more likely to be delinquent. Our results are significant, with an accuracy of up to 81\%. The model developed in our work will be able to help our client attain a better sense of its AR operations and take better actions, thus improving its cash flow. 

Our set of historical features is small and captures the customer behavior payment using temporal information to make better prediction. We demonstrated by our experiments that using the window size with a small number of months (4) we were able to deal with concept drift in the dataset. We also created a new prioritization list that is able to rank customers in a more realistic way, helping the client to optimize their resources with respect to daily action of the collectors. 

As real world environments are always in continuous flux, our features distribution are shifting as well. We noted that the current AR process has been modified over the last year, and, as it evolves, our model should accompany that. We thus recommend a continuous evaluation of the analytics results so as to keep track of accuracy metrics as well as AUC, as discussed in the results section. From time to time, it seems necessary to retrain the model since all processes suffer from what is called concept drift, i.e., the relationship between the features and the labels evolve over time and classical machine learning approaches consider only stationary data.

Finally, we validated our proposed ranking function with a simulation that shows savings up to $\approx$1.75 million dollars, suggesting the proposed approach will have positive impact in the real world.